\documentclass[10pt,letterpaper]{article}
\usepackage{hyperref}
\usepackage{cogsci}

\cogscifinalcopy 

\usepackage{pslatex}
\usepackage{apacite}
\usepackage{float} 
\usepackage{graphicx}
\usepackage{tabularx}
\usepackage{bigdelim, rotating}

\usepackage{gb4e}
\noautomath
\usepackage[dvipsnames]{xcolor}
\usepackage{multirow,booktabs}

\title{Overinformative Question Answering by Humans and Machines}

\author{{\large \bf Polina Tsvilodub (polina.tsvilodub@uni-tuebingen.de)} \\
        {\large \bf Michael Franke (michael.franke@uni-tuebingen.de)} \\
	Department of Linguistics, 
  University of T\"ubingen \\
 \AND
 \begin{tabular}{cc}
      {\large \bf Robert Hawkins (rdhawkins@princeton.edu)} 
      & {\large \bf Noah D. Goodman (ngoodman@stanford.edu)} \\
    Princeton Neuroscience Institute & 
    Departments of Psychology and Computer Science \\ 
    Department of Psychology, Princeton University &
    Stanford University
 \end{tabular}
    }

\begin{document}
	
\maketitle

	\begin{abstract}
		When faced with a polar question, speakers often provide overinformative answers going beyond a simple ``yes'' or ``no''. But what principles guide the selection of additional information? In this paper, we provide experimental evidence from two studies suggesting that overinformativeness in human answering is driven by considerations of \emph{relevance} to the questioner's goals which they flexibly adjust given the functional context in which the question is uttered. We take these human results as a strong benchmark for investigating question-answering performance in state-of-the-art neural language models, conducting an extensive evaluation on items from human experiments. We find that most models fail to adjust their answering behavior in a human-like way and tend to include irrelevant information. We show that GPT-3 is highly sensitive to the form of the prompt and only achieves human-like answer patterns when guided by an example and cognitively-motivated explanation. 
		
	\end{abstract}

\section{Introduction}
Human interlocutors effortlessly select \textit{relevant} information from an abundance of contextually available details \cite{roberts2012information, sperber1986relevance}. 
Understanding the principles underlying these processes in humans has become increasingly critical for building artificial agents that are able to help users interface with large bodies of information, such as rich knowledge bases \cite{thoppilan2022lamda}. 
Human question answering behavior provides a particularly useful window into the problem of relevance. 
For example, imagine that you are a bartender at a caf\'e. Today your caf\'e only serves iced coffee, soda and Chardonnay. Imagine a customer asking: ``Do you have iced tea?'' A natural answer might be, ``I'm sorry, we don't have iced tea, but we have iced coffee!'' 
That is, even though a simple ``no'' is a valid response \cite{hamblin1976questions, karttunen1977syntax} to a polar question, it feels natural to provide additional information for the questioner \cite{clark1979responding, hakulinen2001minimal}. 

But what, exactly, guides the selection of what the respondent chooses to include among many possible details not strictly required by the question (i.e., mentioning the iced coffee, but not the wine)?
One influential theoretical framework has been provided by \citeA{RobertRooijQuestioningDecisionProblems2003} and \citeA{benz2006utility}, who formulate the problem of relevant answers in decision-theoretic terms, suggesting that good answers are ones that resolve the questioner's decision problem. 
A related model of question-answering has been formulated within the Rational Speech Act framework \cite{frank2012predicting} which defines a pragmatic answerer who attempts to infer the questioner's decision problem  \cite{hawkins2015you}. 
While these models address thorny theoretical problems about the semantics and pragmatics of questions, they have been challenging to deploy at scale or evaluate empirically.

An alternative framework for question answering has arisen from the more engineering-oriented natural-language processing (NLP) literature. 
For example, classical approaches aimed to learn similarity-based answering heuristics \cite{quarteroni_manandhar_2009} and  more recent multi-modal approaches aimed to align the compositional structure of the question with the available visual information \cite{andreas2016neural} or to jointly embed visual and textual information \cite{zhou2020unified}. 
Another approach has focused on breakthroughs in \textit{large language models} (LLMs) trained end-to-end on generic language modeling.
These models have achieved human-level performance on  downstream tasks like factual question answering zero-shot, i.e., without being fine-tuned for that specific task \cite{radford2019language, sanh2021multitask}. 

Attempts to understand the mechanisms of these models have revealed intriguing differences between LLM performance and humans on cognitive tasks commonly used in psychology \cite{binz2022using}, as well as the sensitivity of these models to the structure of their prompts \cite{lampinen2022can, lampinen2022context}. 
Yet work in this literature has largely focused on \textit{factual} question-answering rather than the kind of common-sense and context-sensitive scenarios where relevance becomes key.
Here, we seek to understand empirical patterns in human responses to contextualized polar questions, with an eye towards better evaluating neural language models for relevant question answering.
\begin{figure*}[ht]
\centering
    \includegraphics[scale = 0.28]{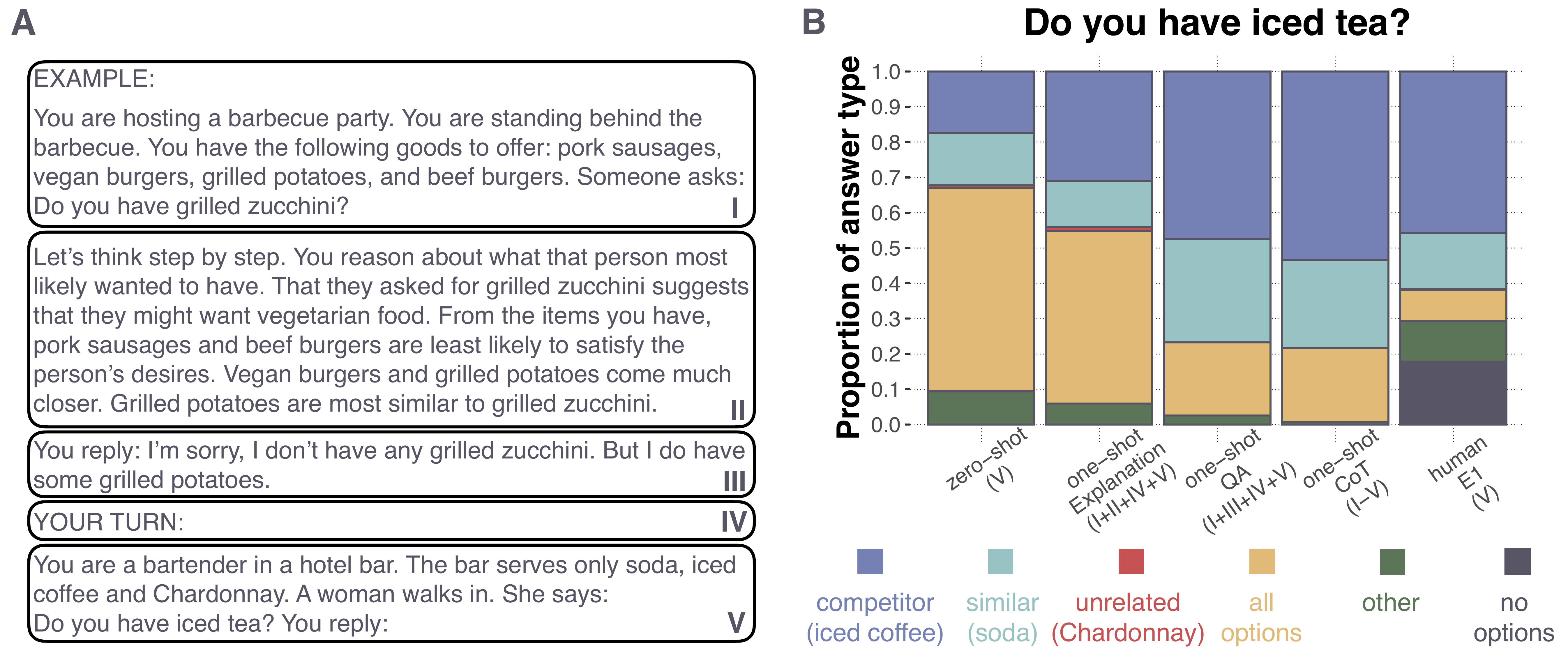}
    \caption{(A) Example prompt from Experiment 1. The target vignette is in block V (bottom). Text blocks with different Roman numbers indicate which text spans were prepended to the target vignette for different one-shot GPT-3 prompts. The one-shot prompts were the same across vignettes. (B) Proportions of different response categories (y-axis) across vignettes for human and GPT-3 samples under different prompts (x-axis). Roman numbers in brackets indicate which text blocks were part of the prompt.
    \label{fig:e1-props}}
\end{figure*}
\section{Experiments}
We hypothesize that human speakers reason about the decision problem the questioner might be facing when asking the question \cite{RobertRooijQuestioningDecisionProblems2003}. Given a detailed context and rich world knowledge, speakers then will act cooperatively and select contextual aspects that they consider relevant for the questioner's problem. More specifically, we hypothesize that, given a polar question in context, humans provide \textit{overinformative} answers and do so by providing additional information that might be relevant for the questioner's practical decision making, but not other information.
We explore this hypothesis by conducting experiments wherein human and artificial agents answered polar questions of a questioner asking if an item was available (e.g., ``Do you have iced tea?''), when embedded in a verbally described context (e.g.,~Fig.~\ref{fig:e1-props}A, Fig.~\ref{fig:e2-props}A). We investigate if and what kind of contextual information agents include in their responses. 

This question was operationalized by manipulating the context preceding a polar question.
The context listed possible options that could be suggested; the options (e.g., iced coffee, soda, Chardonnay) did not include the requested target item (e.g., iced tea). The alternatives included an option that we categorized as the optimal alternative to the requested target for the questioner's practical problem in the given context. We call this alternative the \textit{competitor} option (e.g., iced coffee). A \textit{similar option} (which was conceptually similar to the target but less relevant for the practical problem) was also included in the list (e.g., soda). Finally, the list included an \textit{unrelated option} which we hypothesized to be irrelevant for the uttered request (e.g., Chardonnay).\footnote{This intuitive classification by the authors was corroborated by pilot studies for Experiment 1 wherein participants rated the likability of each alternative, assuming that the target was requested. The results aligned well with the manual classification.} The order of alternatives in the list was randomized within-participant. Copy-pasting text from the context was disabled.

\textit{Overinformative} answers to a polar question like ``Do you have iced tea?'' are answers which go beyond a simple ``no'' answer (what we call here the \textit{no options} response) and include a suggestion of an alternative option or some solution to the request.
There are different kinds of overinfomative answers in the experimental context we consider.
A \textit{competitor} answer offers only the competitor alternative to the questioner (e.g., ``I'm sorry, we don't have iced tea, but we have iced coffee.''). \textit{Similar option} answers mention either the respective option alone, or combined with the competitor (``We have soda (and iced coffee).''). 
\textit{Unrelated option} answers mention the respective option, too (``We have Chardonnay.''), while \textit{all options} answers enumerate all available alternatives (``We have iced coffee, soda and Chardonnay.''). 
Details added to this set up in the single experiments are described below.\footnote{All experimental and supplementary materials, data and analyses can be found under \url{https://tinyurl.com/42krksx7}}
\subsection{Experiment 1: Eliciting Overinformativity}
The first experiment elicited free production responses to polar questions given contexts which were relatively uninformative with respect to the questioner's action goals, i.e., participants had to make inferences about the likely intended goal based on the question itself. If participants are overinformative, we expect a preference for generating \textit{competitor}, \textit{similar option} or \textit{all options} responses over providing a \textit{no options} answer.
Further, if human overinformativity is based on reasoning about the relevance of the alternative, we expect a preference towards \textit{competitor} and \textit{similar option} responses over \textit{unrelated option} or \textit{all options} responses.

\subsubsection{Participants} We recruited 162 participants via the crowd-sourcing platform Prolific. Four participants were excluded due to failing a simple attention check wherein they were instructed to type a particular word in the free response box. Both experiments took around four minutes and participants were reimbursed with \pounds0.60. Participants in both experiments were restricted to self-reported native speakers of English, located in the US and UK, with Prolific approval rates over 95\% and at least five previously completed Prolific studies.

\subsubsection{Materials and Procedure} In this experiment, we used 15 pairs of vignettes (30 in total; e.g., Fig.~\ref{fig:e1-props}A, last paragraph). In each pair, the options described in the context of a vignette included a \textit{competitor} (e.g., iced coffee), a \textit{similar option} (e.g., soda), and an \textit{unrelated option} (e.g., Chardonnay). The pairs were constructed such that, e.g., iced coffee was the \emph{competitor} in one vignette and the \emph{unrelated} option in the second vignette of a pair, and vice versa for, e.g., Chardonnay. This was done to tease apart possible frequency effects of the options from context-based reasoning for the LLM experiments. 
Each participant saw four vignettes sampled at random, each consisting of one question. The four main trials were shuffled with one attention check. On each trial, they read the background context followed by the question. They freely typed their answer following the prompt ``You reply:''. Participants were instructed to type responses to the question they read, given the vignette, in a harmless and helpful manner. 
\subsubsection{Results} 
Responses stating falsely that the requested item was available were excluded (4\%).
Remaining responses were manually classified into the following response types described above: \textit{competitor}, \textit{similar option}, \textit{unrelated option}, \textit{all options}, \textit{no options} and \textit{other} responses. The last category contained responses providing an answer to the question without suggesting an explicit alternative from the context but providing an alternative solution like a reason for the unavailability, information about alternatives in more general terms, follow-up questions and comments, as well as suggesting both the related and unrelated options. 

Consistent with our expectations, participants preferred competitor responses over all other response types (Fig.~\ref{fig:e1-props}B, dark blue bars). Furthermore, they preferred suggesting only alternatives similar to the target over suggesting unrelated options or enumerating all available alternatives. 

These differences are corroborated by a Bayesian multinomial regression model, regressing the response type against an intercept, with the competitor response category coded as the reference level (mean and 95\% credible intervals reported for both experiments): participants preferred competitor responses over similar-option responses ($\beta=-1.06~[-1.29, -0.83]$), competitor responses over unrelated-option and all-options responses ($\beta=-4.93~[-6.60, -3.73]$ and $\beta=-1.65~[-1.96, -1.36]$), and similar-option responses over unrelated-option and all-options responses ($\beta=0.59~[0.59, 0.93]$ and $\beta=3.87~[2.61, 5.55]$). They also credibly preferred offering an alternative over providing a no-option response ($\beta=-1.52~[-1.74, -1.32]$), as well as providing competitor, similar-option and unrelated-option responses over all-options responses ($\beta = -1.95~[-2.26, -1.67]$).
\subsection{Experiment 2: Manipulating Functional Context}
We hypothesize that the driving force behind the speakers' selection of the additional information in their responses is the reasoning about the questioners' action goals. Therefore, we designed a second experiment to more clearly distinguish action relevance from similarity of alternative options. 
That is, in this experiment we manipulated the relevance of the alternatives by creating vignettes which contained a list of options, but presented the same alternatives in two different contexts suggesting different motivations (Fig.~\ref{fig:e2-props}A), hence rendering distinct options more relevant for the questioner. If our hypothesis is true, we would expect that participants provide different options in their overinformative responses given each of the contexts belonging to one vignette pair. 

\subsubsection{Participants} We recruited 130 participants via Prolific. Ten participants were excluded due to failing an attention check identical to Experiment~1, and one due to only providing (infelicitous) positive responses. 

\subsubsection{Materials and Procedure} We designed 12 vignettes, consisting of pairs of contexts which both included the same alternative options but described different action problems the questioners faced (Fig.~\ref{fig:e2-props}A). Crucially, the alternatives included an option a priori \textit{most similar} to the requested target, two distinct \textit{competitors} anticipated to be optimal for a respective context, and an \textit{unrelated option} irrelevant for both contexts. For each context, the option that was anticipated to be the competitor was treated as a \textit{similar option} for the second context. Each participant saw only one of the two possible contexts per vignette. Four vignettes were sampled at random per participant and an attention check was added. Participants received identical instructions to Experiment~1. 

\subsubsection{Results}
False responses were excluded (3\%).
Remaining responses were manually classified into the same categories as in Experiment~1, with the addition of the \textit{most similar} response type which included responses mentioning the a priori \textit{most similar} option. The \textit{similar option} category included responses mentioning subsets of the \textit{competitor, most similar, similar} options. The \textit{other} category also included responses offering both the \textit{most similar} and the \textit{unrelated} options. Additionally, each response that mentioned alternative options was annotated with the type of the mentioned options (\textit{competitor 1} for context 1, \textit{competitor 2} for context 2, a priori \textit{most similar} option, and \textit{unrelated option}). The context numbering indicates distinction, not a meaningful numbering.
\begin{figure*}[ht]
\centering
    \includegraphics[scale = 0.26]{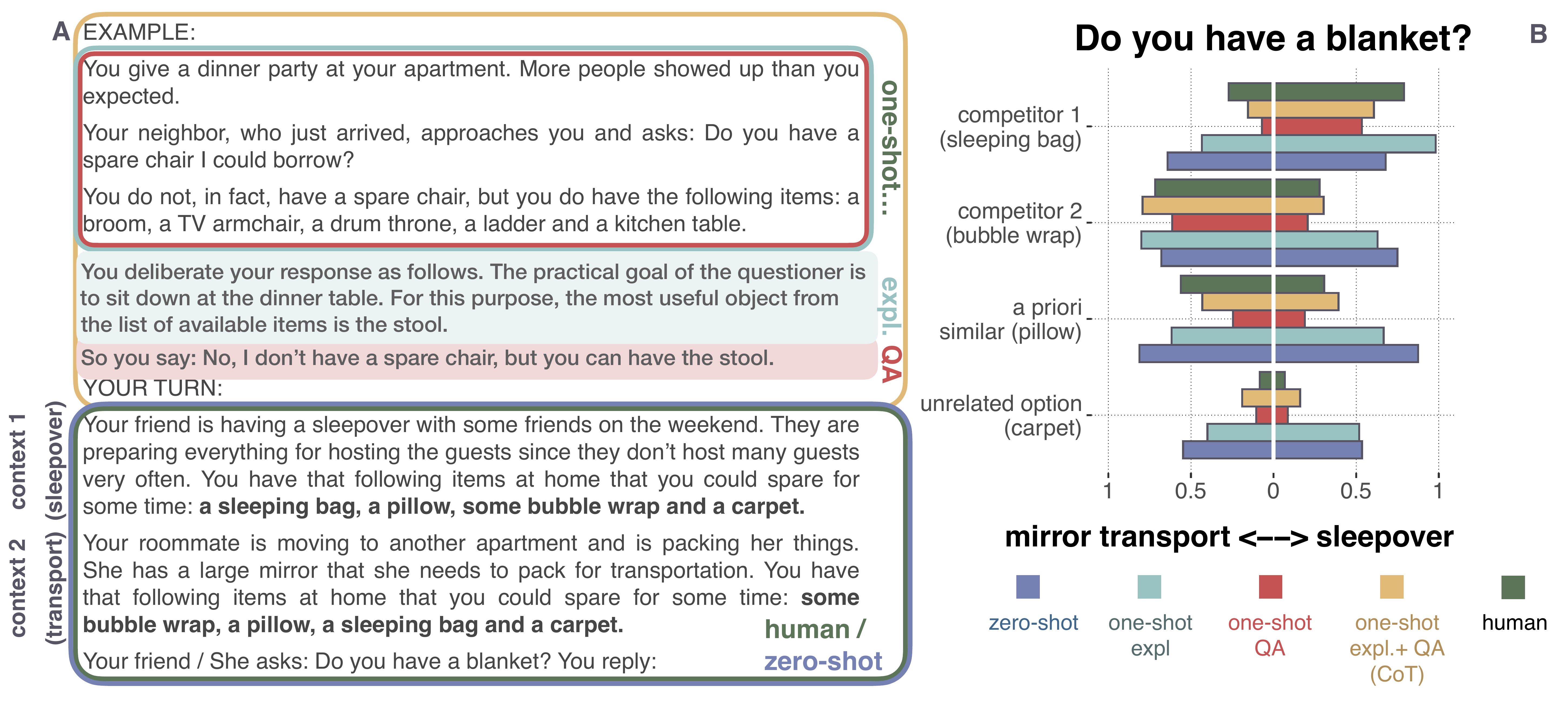} 
    \caption{A: Example vignettes from Experiment 2. Both contexts possible for one vignette are shown (last box, context 1~vs.~context 2). The colored boxes indicate which parts of the texts were used for the human experiment and different GPT-3 prompts. The example vignette is marked dark blue and dark green (bottom). B: Mentioning proportions of different alternatives in all responses (y-axis) in each context of a vignette (mirrored on x-axis) across vignettes in Experiment 2. Human results are plotted against GPT-3 samples given different prompts (colors). \label{fig:e2-props}} 
\end{figure*}

\begin{table*}[ht]
\begin{center} 
\caption{Response type proportions for humans and GPT-3 in different prompting conditions across contexts in Experiment 2.} 
\label{tab:e2} 
\vskip 0.12in
\begin{tabularx}{\textwidth}{lllllllll} 
\hline
\textbf{Category} &  Competitor & Most similar & Similar & Unrelated & All & Other & No options\\
\hline
\textbf{Human} & 0.33 & 0.10 & 0.33 & 0.00 & 0.03 & 0.13 & 0.08	\\
\textbf{GPT-3 zero-shot} & 0.06 & 0.09 & 0.31 & 0.00 & 0.42 & 0.12 & 0.00\\
\textbf{GPT-3 one-shot explanation} & 0.28 & 0.04 & 0.21 & 0.03 & 0.39 & 0.04 & 0.00\\
\textbf{GPT-3 one-shot QA} & 0.54 & 0.18 & 0.15 & 0.08 & 0.00 & 0.05 & 0.00\\
\textbf{GPT-3 one-shot CoT} & 0.45 & 0.14 & 0.22 & 0.05 & 0.09 & 0.04 & 0.00\\
\hline
\end{tabularx} 
\end{center} 
\end{table*}
Similarly to Experiment 1, we found that participants were overinformative and mentioned relevant alternative options: compared to producing \textit{competitor} responses, they were less likely to produce \textit{all options} ($\beta=-2.47~[-3.09, -1.91]$), \textit{most similar} ($\beta=-1.22~[-1.57, -0.88$]), \textit{unrelated} ($\beta=-4.28~[-5.93, -3.08]$), \textit{other} ($\beta=-0.91~[-1.22, -0.62]$) or \textit{no options} responses ($\beta=-1.43~[-1.79, -1.07]$). Participants also preferred providing \textit{similar options} responses over \textit{all options} ($\beta=2.56~[1.99,3.20]$), \textit{most similar} ($\beta=1.28~[0.95, 1.62]$), \textit{unrelated} ($\beta=4.35~[3.16,5.94]$) and \textit{no options} responses ($\beta=1.48~[1.14,1.86]$).
Furthermore, confirming our expectations, participants offered different relevant alternatives in one context compared to the other for a given vignette (Fig.~\ref{fig:e2-props}B, green bars, left half~vs.~right half). This was confirmed by a multinomial Bayesian regression model, fitting an intercept, a fixed effect of context and a random by-vignette intercept to the type of mentioned option (relative to context 1 competitor): participants were more likely to mention competitors relevant for context 1 in context 1 than in context 2 ($\beta=19.35~[11.92, 29.81]$), more likely to mention competitors relevant for context 2 in context 2 than in context 1 ($\beta=3.45~[2.46, 4.45]$), more likely to mention most similar and unrelated options in context 2 than to mention the respective option in context 1 ($\beta=2.79~[1.90, 3.71]$, $\beta=3.91~[2.45, 5.53]$). Taken together, these results indicate that the options the speakers considered functionally relevant in the respective context differed.  

\section{Neural Language Model Analysis}
To evaluate the extent to which state-of-the-art models are capable of human-like context-sensitive overinformative question answering, we test the performance of various neural models on the same vignettes and questions from human experiments. We provide a particularly detailed performance analysis of GPT-3, a large language model which has been shown to achieve strikingly human-like results on various NLG tasks \cite{sanh2021multitask}. If neural models are capable of human-like overinformative question answering, we expect the same pattern of response types as in human experiments.
\subsubsection{Participants} Two different types of neural models were evaluated. 
The first type were the following \textit{extractive} models which were fine-tuned with a question answering head on common question-answering datasets: RoBERTa, BERT L, BERT base, DistilBERT (cased, uncased), DeBERTa, tinyRoBERTa, Electra, BART \cite{devlin2018bert, he2020deberta, clark2020electra, lewis2019bart, liu2019roberta, sanh2019distilbert}. These models predict the position of a span in the input text which contains the predicted answer to the question. 
The second type were the following \textit{generative}, or, causal language models (LMs) which are pretrained on the language modeling objective: T0, GPT-2, GPT-3 davinci-003 and ChatGPT \cite{radford2019language, NEURIPS2020_1457c0d6, sanh2021multitask, chatgpt}. Versions of BART and T5 fine-tuned for free answer generation were  also tested \cite{lewis2019bart, raffel2020exploring}. 
All models used in this study were not fine-tuned to this task.\footnote{See repository above for details on datasets and model access.} 

\subsubsection{Materials and Procedure}
All vignettes from experiments 1 and 2  were used. To optimize the conditions in favor of the models, the competitor option always occurred as the first alternative in the list of options in context. We sampled the top five predicted responses to the questions, conditioning on the contexts (for ChatGPT only one response was retrieved). For LMs, these were retrieved by using beam search with beam size five, sampling temperature 1 and maximal prediction length of 64 tokens. For extractive models, these were the five highest probability spans of length $>0$.

Furthermore, building on previous studies that investigated the effects of prompting  GPT-3 \cite{lampinen2022context}, we compared off-the-shelf \textit{zero-shot} predictions of GPT-3 to \textit{one-shot} predictions wherein the context was preceded by an example of the task. The \textit{one-shot QA} prompt contained an example context, question and the expected competitor response to that question; the \textit{one-shot chain-of-thought} (CoT) prompt additionally provided an explanation of the relevance-based reasoning for the selection of the competitor before the example competitor response (Fig.~1A, ``expl.''); the \textit{one-shot explanation} prompt contained an example context, question and the CoT, but no example of the respective response (see Fig.~\ref{fig:e1-props}A, \ref{fig:e2-props}A), allowing to tease apart the importance of the example reasoning in contrast to an example of an answer offering only one (most relevant) alternative.\footnote{For Experiment 1, various versions of the prompts were tested. A prompt including ``Let's \{think, reason\} step by step'' following ``YOUR TURN'' did not change results qualitatively. In one-shot explanation prompts, it led to some responses actually spelling out the same reasoning, without a response prediction. Presented one-shot explanation results were compared to results with the prompt containing ``You respond accordingly.'' after the CoT. No qualitative differences were observed.} Exploratory tests revealed that one-shot prompting of other LMs often led to nonsense performance so it is omitted. Only the CoT prompt is used for the ChatGPT one-shot condition.

The sampled responses were manually classified following the same annotation scheme as for human responses. For Experiment 2, as before, the predictions were additionally annotated with respect to the option types they mentioned.

Additionally, we computed the log probabilities of different response types predicted by the LMs. To this end, we constructed long responses of the form ``I'm sorry, \{I, we\} don't have X. \{I, We\} have Y'' where X was the target and Y a set of alternatives defining the different response types (see above). Probabilities over sentences with all permutations of options in Y were averaged. Long response scores were averaged with short responses omitting the first sentence. Retrieved probabilities were renormalized over the response types.

\subsubsection{Results for Experiment 1} 
The modal response type differed across extractive models.
For models based on BERT, the \textit{competitor} proportions came close to human data (see supplementary materials repository for all results other than GPT-3 from both experiments).
However, the second most popular response type was shared between \textit{all}, \textit{no options} or \textit{other} responses across models, while humans preferred \textit{similar} option responses. We observed a high proportion of \textit{other} responses due to nonsense spans containing context or mixed-category options.

For generative models, we found that all models besides GPT-3 and one-shot chatGPT were much less likely to predict \textit{competitor} or \textit{similar} compared to \textit{all options} or \textit{other} responses. 
Only GPT-2 and T0 also produced a noticeable proportion of \textit{no options} responses. 
The probabilities of different response types generated by language models also differed from human response proportions in that they assigned more uniform probabilities to the different options (i.e., they did not mirror any clear preferences for particular response types).
The computed probabilities also differed from the response type proportions sampled from the models themselves, indicating that probabilities generated by LMs might not be representative of their free inference behavior. 
Human response proportions differed significantly from the response proportions of all models ($\chi^2$-test, all $p$-values $<0.05$).

For GPT-3 and ChatGPT, we found that the models were most likely to produce \textit{all options} responses in the zero-shot and the one-shot explanation settings, and matched human response patterns more closely by producing predominantly \textit{competitor} responses when given an example of the competitor response (i.e., with the one-shot QA and CoT prompts; Fig.~\ref{fig:e1-props}B).\footnote{Samples retrieved through the GPT-3 API often contained answers consisting of the end-of-sequence token (EOS) only (0.1-0.4 of samples). These were excluded from analyses, response category proportions were renormalized over remaining dataset. Results for one-shot explanation prompting in Figure~\ref{fig:e1-props} (light blue bars) are averaged over two sets of samples. The absence of no options responses is a feature of the GPT-3 davinci-003 model version. GPT-3 davinci-002 produced up 0.5 of responses of no options type, depending on the prompting condition.} Yet even given the one-shot QA and CoT prompts, GPT-3 and ChatGPT were more likely to generate \textit{all options} responses than humans. ChatGPT was slightly less likely to produce \textit{competitor} responses in the one-shot CoT condition compared to GPT-3.
Human response patterns significantly differed from GPT-3 responses in all prompting conditions ($\chi^2$-test, all $p$-values $<0.05$). We conclude that, given the right prompting, the models were sensitive to the context (as opposed to frequency of the alternatives, because the models produced different responses across pairs).
Overall, these results indicate that GPT-3 is strikingly sensitive to prompting even with a single example, but including an example of the optimal response is more crucial for achieving human-like response patterns than explaining the underlying reasoning. 
	
\subsubsection{Results for Experiment 2}
Similarly to Experiment 1, different extractive models showed different patterns of preferences over response types. 
None of the models came close to predicting the human preference for contextualized \textit{competitor} responses.
In contrast to Experiment 1, BERT-based models mostly generated \textit{other} responses which might have contained both similar and unrelated alternatives.

Turning to generative models other than GPT-3, we found that no model came close to matching the human preference for \textit{competitor} responses (see repository).
Instead, the models showed different patterns, often producing \textit{all}, \textit{no} or \textit{most similar} and \textit{other} responses. Indeed, human response proportions and model response proportions differed significantly ($\chi^2$-test, all $p$-values $<0.05$).
This indicates that in a context which requires reasoning about functional relevance both generative and extractive models are further from human performance, compared to a setting where the competitor might be selected via a general similarity metric.

To assess the performance of GPT-3 in more detail, two ways of looking at its predictions are useful.
\autoref{tab:e2} shows proportions of answer types using similar categories as for Experiment~1.
\autoref{fig:e2-props}B considers answer components independently for each context, zooming in on the context-dependence of answers.
Both analyses of GPT-3's performance show considerable variance under different prompts.\footnote{
  For GPT-3, we found that a prompt containing an example of reasoning about \textit{functional} relevance was critical for predictions to match human results.
  That is, exploratory studies revealed that GPT-3 was more likely to offer \textit{all options, similar options} or the a priori \textit{most similar} alternative compared to humans when prompted with the same prompts as in Experiment 1.
  ChatGPT showed a similar pattern (see repository).
} 

Looking at \autoref{tab:e2}, we find a large proportion of \textit{all options} responses in the zero-shot condition for GPT-3, as in Experiment~1.
The \textit{all options} proportion was also relatively high in the one-shot explanation condition for GPT-3, but the model was also able to produce more \textit{competitor} responses, indicating that an explanation involving contextual relevance might bias the model towards selecting the appropriate competitor.
Both in the one-shot QA and one-shot CoT conditions, GPT-3 even outperformed humans by producing more contextually-appropriate \textit{competitor} responses and less \textit{similar} and \textit{other} responses. 
In contrast to Experiment 1, the proportion of \textit{competitor} responses was larger in the one-shot QA condition than in the one-shot CoT condition. 
In both conditions, GPT-3 still produced more \textit{most similar} responses than humans. GPT-3 produced more exhaustive \textit{all options} responses except with the one-shot QA prompt.
For all prompting conditions, human response proportions differed significantly from model response proportions ($\chi^2$-test, all $p$-values $<0.05$).

Different performance for different types of prompts shows even more strikingly, when we consider the ability to single out the competitor as the most relevant option in different contexts.
\autoref{fig:e2-props}B (left~vs.~right half) shows the overall frequency with which particular items were mentioned in responses, separated for the two contexts of each vignette.
Only in the QA and CoT prompts was the model able to flexibly identify the functionally appropriate competitor in context.
With these prompts it also produced less irrelevant options (\textit{most similar} and \textit{unrelated}) than for other prompts.
Finally, comparing the QA and CoT prompts, GPT-3 was less likely to include these irrelevant options given QA than given CoT (\autoref{fig:e2-props}B, red~vs.~yellow bars).

Taken together, these results indicate that GPT-3 cannot spontaneously identify the ``relevant relevance dimension'' for a question.
But once prompted with the appropriate dimension, it is strikingly sensitive to examples that include the intended form of the answer, even outperforming humans by producing strictly more informative answers.

\section{Discussion}
Taken together, our results in Experiment 1 provide evidence in favor of the hypothesis that human overinformativity is driven by reasoning about relevance based on the observed question. Moreover, Experiment 2 revealed that humans adjust what they consider to be relevant depending on context, in particular, for the functional problem of the questioner.

Our comparison of human data to samples from SOTA neural models revealed that these often fail to select the relevant subset of contextually available options and instead include all information.
The analysis of GPT-3's performance, when it is conditioned on different prompts, is in line with previous research, showing that prompting form matters \cite{binz2022using}.
Specifically, GPT-3 came closest to or outperformed human behavior in terms of only providing the most necessary information when it was prompted with an example target response, but not given a prompt describing the underlying reasoning only. 

These results open up several avenues for further research. 
The language model results were obtained using the suggested default decoding scheme parameters. 
Yet the decoding scheme is known to affect results \cite{holtzman2019curious, meister-etal-2020-beam}, so that investigating interactions of decoding schemes and different prompts may be insightful.
We further observed qualitative differences in the types of responses preferred by language models in terms of samples and probabilities assigned to given response types. This suggests that the latter analysis might rather reflect the models’ surface form preferences of the scored sentences than qualitative reasoning characteristics.
Finally, the difference in propensity to generate \textit{all options} responses between humans and neural models might be due to humans being more sensitive to production effort than machines.
Since copying-pasting from the context was disabled for human participants, adding a production length penalty to neural models might bring their response patterns closer to human-like preferences.

In sum, we have presented novel empirical evidence that human question answering is guided by subtle, context-sensitive pragmatic reasoning mechanisms, and we argued that these abilities should not be taken for granted even for very sophisticated neural language models.
These capacities may be implicit in neural models fine-tuned on user response data, but require lucky prompt-engineering and hand-holding to be fully expressed.
\newpage
 
\section{Acknowledgments}
Michael Franke is a member of the Machine Learning Cluster of Excellence, EXC number 2064/1 – Project number 39072764.
\bibliographystyle{apacite}
	
\setlength{\bibleftmargin}{.125in}
\setlength{\bibindent}{-\bibleftmargin}
	
\bibliography{CogSci_Template}

\end{document}